\documentclass{article}
\PassOptionsToPackage{numbers, compress}{natbib}

\usepackage[final]{neurips_data_2024}
\usepackage{amsmath}
\pdfoutput=1

\usepackage{multirow}
\usepackage[utf8]{inputenc} 
\usepackage[T1]{fontenc}    
\usepackage{url}            
\usepackage{booktabs}       
\usepackage{amsfonts}       
\usepackage{nicefrac}       
\usepackage{microtype}      
\usepackage{xcolor}         
\definecolor{cvprblue}{rgb}{0.21,0.49,0.74}
\definecolor{txblue}{rgb}{0,1,1}
\usepackage[breaklinks=true,colorlinks=true,citecolor=cvprblue,bookmarks=false]{hyperref}
\usepackage{colortbl}  
\usepackage{enumitem}
\usepackage{graphicx}

\usepackage{subcaption}
\usepackage{cleveref}   
\def\ourdataset{OpenSatMap}

\usepackage{wrapfig}

\title{\ourdataset: A Fine-grained High-resolution Satellite Dataset for Large-scale Map Construction}
\usepackage{mathrsfs}
\newlength\savewidth\newcommand\shline{\noalign{\global\savewidth\arrayrulewidth
  \global\arrayrulewidth 1pt}\hline\noalign{\global\arrayrulewidth\savewidth}}

\Crefname{figure}{Fig.}{Figs.}
\Crefname{section}{Sec.}{Secs.}
\Crefname{table}{Tab.}{Tabs.}

\author{%
  Hongbo Zhao$^{1,3}$\thanks{Equal contribution.} 
    \quad Lue Fan$^{1,3*\dag}$
    \quad Yuntao Chen$^{2*}$
    \quad Haochen Wang$^{1,3*}$ 
    \quad Yuran Yang$^{4,5*}$\\
    \textbf{Xiaojuan Jin}$^{1}$  
    \quad \textbf{Yixin Zhang}$^{4}$  
    \quad \textbf{Gaofeng Meng}$^{1,2,3}$   
    \quad \textbf{Zhaoxiang Zhang}$^{1,2,3}\thanks{Corresponding author.}$ \\
    $^1$Institute of Automation, Chinese Academy of Sciences (CASIA) \\
    $^2$Centre for Artificial Intelligence and Robotics, HKISI, CAS  \\
    $^3$University of Chinese Academy of Sciences (UCAS)  \\
    $^4$Tencent Maps, Tencent \, $^5$Beijing University of Posts and Telecommunications \\
{\url{https://opensatmap.github.io}}
}

\begin{document}

\maketitle

\begin{figure}[h]
    \vspace{-25pt}
    \centering
    \includegraphics[width=\linewidth]{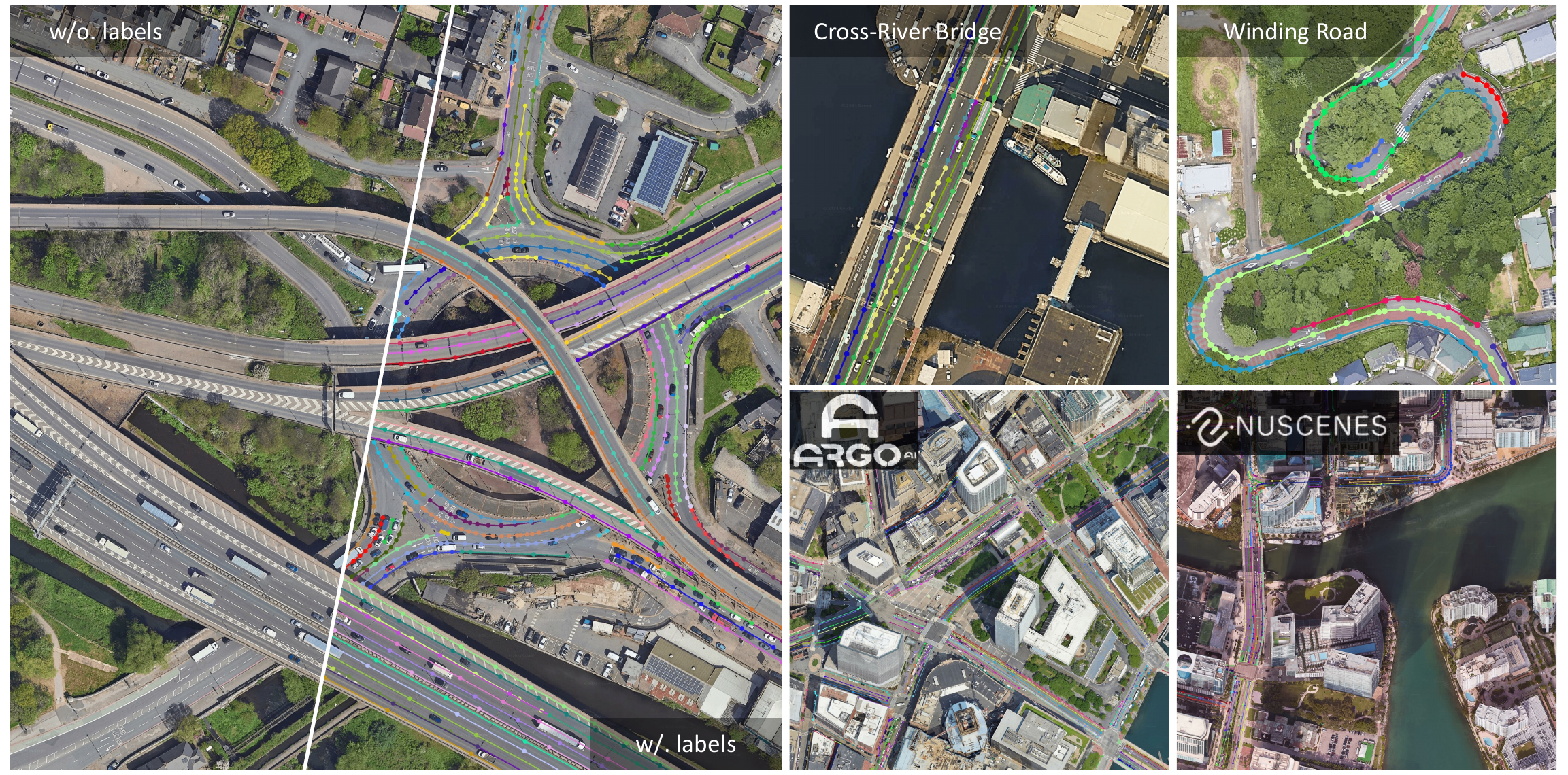}
    \caption{
    \textbf{Demonstrations of \ourdataset{} dataset}. It contains high-resolution satellite images with fine-grained annotations, covering diverse geographic locations and popular driving datasets~\cite{caesar2020nuscenes,chang2019argoverse}. 
    }
    \label{fig:fig1}
\end{figure}

\begin{abstract}
In this paper, we propose \ourdataset, a fine-grained, high-resolution satellite dataset for large-scale map construction.
Map construction is one of the foundations of the transportation industry, such as navigation and autonomous driving.
Extracting road structures from satellite images is an efficient way to construct large-scale maps.
However, existing satellite datasets provide only coarse semantic-level labels with a relatively low resolution (up to level 19), impeding the advancement of this field.
In contrast, the proposed \ourdataset{} (1) has fine-grained instance-level annotations; (2) consists of high-resolution images (level 20); (3) is currently the largest one of its kind; (4) collects data with high diversity.
Moreover, \ourdataset{} covers and aligns with the popular nuScenes dataset and Argoverse 2 dataset to potentially advance autonomous driving technologies.
By publishing and maintaining the dataset, we provide a high-quality benchmark for satellite-based map construction and downstream tasks like autonomous driving.
\end{abstract}

\section{Introduction}
\label{intro}

Large-scale, up-to-date, and fine-grained map construction is fundamental to many tasks such as traffic control and autonomous driving.
Compared with on-road mapping, constructing maps from satellite images is a highly efficient way for this purpose because of the large geographic coverage.

Although it is promising, parsing fine-grained road structures from satellite images for map construction is a highly challenging task for the following reasons.
\textit{(i)} The resolution of satellite images is usually not high enough for fine-grained lane line detection.
For example, the common level-19 satellite images have a resolution of 30 cm per pixel, which can barely make out a lane line that is 20 cm wide.
\textit{(ii)} The lane lines in modern cities possess highly complex topological structures and function-based semantics.
It not only poses challenges for designing models with enough capacities but also makes it difficult to build fine-grained datasets.

Existing benchmarks \cite{bastani2018roadtracer, cheng2017automatic, demir2018deepglobe, he2020sat2graph, gao2024ddctnet} have made attempts to handle the challenging tasks. 
However, they each have their limitations, preventing them from effectively supporting large-scale and fine-grained map construction.
Specifically, existing benchmarks have the following limitations.
\begin{itemize} \label{limitations}
    \item \textbf{Coarse annotations}. All of these benchmarks support only coarse semantic-level lane segmentation, which is far from enough to parse fine-grained lane markings with various functionalities and highly complicated topologies. 
    \item \textbf{Low resolution}. The images in existing benchmarks often have relatively low resolutions. The highest resolution they have is 0.3 m per pixel, which is inadequate for accurate perception of lane lines.
    \item \textbf{Small scale}. Existing benchmarks have relatively small scales. The largest of them have less than 10k 1024 $\times$ 1024 images.
    \item \textbf{Unalignment with autonomous driving benchmarks}. Existing benchmarks solely focus on specific regions, limiting their further applications in autonomous driving, which also heavily relies on map construction. 
\end{itemize}
Table~\ref{tab:dataset_comp} shows the basic information of them.
Given the limitations of existing benchmarks and the challenges of this task, we construct \textbf{\ourdataset}, a new dataset for map construction from satellite images, with the following unique features.
\begin{itemize}
    \item \textbf{Fine-grained instance-level annotations}. Unlike the coarse semantic level annotations in previous datasets, we carefully label fine-grained instance-level lane lines according to fine-grained line attributes. 
    \item \textbf{Higher resolution.} We collect level-20 satellite images with a resolution of 0.15 m per pixel, which is higher than the resolutions of all previous benchmarks. 
    Considering the data accessibility and diversity, we additionally provide level-19 images with a resolution of 0.3 m per pixel.
    \item \textbf{Larger scale}. We collect and carefully annotate \textbf{38k} 1024 $ \times$ 1024 images and around \textbf{445k} instances, which is around \textbf{5x larger} than the largest one of the previous datasets in terms of the number of images.
    \item \textbf{Higher diversity.} We collect data from 60 cities and 19 countries around the world. The collected data is highly diverse and covers various road types, geographic environments, special structures, and traffic regulations. 
    \item \textbf{Alignment with autonomous driving benchmarks.} In order to advance the map construction in autonomous driving, \ourdataset{} covers the regions of nuScenes \cite{caesar2020nuscenes} and Argoverse 2 \cite{chang2019argoverse} datasets, which are the most popular autonomous driving datasets. In addition, we manually align the GPS locations of our data with them for practical use.
\end{itemize}
Given these characteristics, we believe \ourdataset{} can serve as a foundation for various applications, including city-scale map construction, lane detection, and autonomous driving.

\section{Related Work}
\textbf{Satellite Imagery Datasets.}
Satellite imagery datasets can be used for several computer vision tasks such as instance segmentation \cite{garnot2021panoptic, van2021multi, chiu2020agriculture}, object detection \cite{9560031, Xia_2018_CVPR, ding2018learning, lam2018xview}, semantic segmentation \cite{wang2021loveda, castillo2022semi,rahnemoonfar2021floodnet, liu2018roadnet}, scene classification \cite{sumbul2019bigearthnet, helber2019eurosat}, and change detection \cite{8518015, 9394710, tian2020hi, Chen2020} \textit{etc}..
These datasets include optical and hyper-spectral images with varying resolutions, covering a wide range of application fields such as urban \cite{papadomanolaki2021deep, castillo2022semi}, agricultural \cite{garnot2021panoptic, chiu2020agriculture}, transportation \cite{liu2018roadnet,buchner2023learning,bastani2018roadtracer} and marine \cite{NOAAFish25:online, gasienica2021ensemble} areas on a global scale.  
They provide a wealth of labeling information, such as building outlines, road networks, vegetation types, water distribution, \textit{etc}., which greatly promotes the development of deep learning in the field of remote sensing images.
In this paper, we are interested in the application of satellite images on road extraction and we will discuss the datasets for road construction using remote sensing imagery later.

\textbf{Satellite Imagery Datasets for Road Extraction.}
Early datasets mainly focus on semantic labeling for the satellite images and contribute significantly to semantic-level road extraction.
Massachusetts \cite{MnihThesis} contains 1171 images of size 1500 $\times$ 1500 with a resolution of 1 m/pixel.
DeepGlobe \cite{demir2018deepglobe} collects images from Thailand, Indonesia, and India and scrapes labels from QGIS \cite{QGIS}. It contains 8570 images with image size 1024 $\times$ 1024.
Roadtracer \cite{bastani2018roadtracer} uses OpenStreetMap \cite{OpenStre43:online} to label 300 images from big cities with 0.6 m/pixel.
SpaceNet \cite{van2018spacenet} contains 2780 images collected from Las Vegas, Paris, Shanghai, and Khartoum.
There are some datasets manually labeled.
Ottawa \cite{liu2018roadnet} collects several typical urban areas with 0.30 m/pixel spatial resolution and annotates the road surface, road edges and road centerlines.
CHN6-CUG \cite{zhu2021global} collects 6 cities in China with 0.5 m/pixel.
and CasNet \cite{cheng2017automatic} is composed of 224 images with a spatial resolution of 1.2 m per pixel.

However, existing satellite road extraction datasets tend to be relatively small \cite{liu2018roadnet, cheng2017automatic,he2022lane,zang2020lane}, or labeled by OpenStreetMap \cite{van2018spacenet,MnihThesis, bastani2018roadtracer} and QGIS \cite{demir2018deepglobe,yao2024building} platforms, which are limited by the meta information in the database.
Besides, \textit{semantic} labeling results in a wealth of under-utilized information.
With the booming development of high-resolution remote sensing imagery, 
academics urgently need a higher resolution, more richly labeled dataset.
Our work fills this gap by providing a high quality, higher resolution, and bigger dataset with \textit{instance-level} vectorized
annotations.
The readers may refer to Table~\ref{tab:dataset_comp} for a detailed comparison against previous datasets.

\begin{table}[t]
\centering\small
\setlength{\tabcolsep}{4pt}
\caption{\textbf{Comparison against previous datasets.} \ourdataset~is the largest road extraction dataset with the highest resolution and the most detailed annotations. 
$^*$ denotes that we \textit{standardize the size of images to 1024 $\times$ 1024} and calculate the number of images in all datasets. 
}
\vspace{2mm}
\begin{tabular}{l|ccccc}
\toprule
Dataset & \# of Images$^*$ & Resolution & GT Source & Labeling Level & Region \\
\midrule
Massachusetts \cite{MnihThesis} & 2513  & 1.00 \textit{m/pixel}   & OSM & Semantic & America \\
CasNet \cite{cheng2017automatic} & 77 &   1.20 \textit{m/pixel}  & Manually & Semantic & - \\
DeepGlobe \cite{demir2018deepglobe} & 8570 & 0.50 \textit{m/pixel}  & QGIS & Semantic & 3 Counties\\
SpaceNet \cite{van2018spacenet}  & 4481 & 0.31 \textit{m/pixel}   & OSM &Semantic & 4 Counties\\
Roadtracer \cite{bastani2018roadtracer} & 4800 & 0.60 \textit{m/pixel}  & OSM &Semantic  & 6 Counties\\
Ottawa \cite{liu2018roadnet}  & 235 & 0.30 \textit{m/pixel}   & Manually &Semantic & Canada\\
CHN6-CUG \cite{zhu2021global}& 4511 & 0.50 \textit{m/pixel}  & Manually & Semantic & China\\
\midrule
\multirow{2}{*}{\ourdataset~(Ours)}  & 7224 & 0.30 \textit{m/pixel} & \multirow{2}{*}{Manually} & \multirow{2}{*}{\begin{tabular}[c]{@{}l@{}} \ Instance,\\ Vectorized\end{tabular}} &  \multirow{2}{*}{\begin{tabular}[c]{@{}l@{}}\quad60 Cities,\\ 19 Countries\end{tabular}}\\
  & 31696 & 0.15 \textit{m/pixel}   \\
\bottomrule
\end{tabular}
\label{tab:dataset_comp}
\end{table}

\section{\ourdataset~Dataset}

In this section, we introduce our pipeline to collect (Sec. \ref{sec:collection}), annotate (Sec. \ref{sec:annotation}) high-resolution satellite images and demonstrate the statistics of \ourdataset{} in Sec. \ref{sec:stats}.

\subsection{Data Collection} \label{sec:collection}

We collect images from Google Maps \cite{GoogleMap} using public Maps Static API, which is publicly available and has been processed by Google to remove sensitive information.
Considering the limitations mentioned in \Cref{intro}, we collect the data \ourdataset{} with the following guidelines and factors.

\textbf{High resolution.} 
To fulfill the goal of higher resolution. We collect level-20 satellite images, which have a resolution of 0.15 cm/pixel.
In this way, our dataset has a higher resolution than all the existing datasets as Table~\ref{tab:dataset_comp} shows.
Google Maps does not have real level-20 images in some regions. Therefore, during the collection, we carefully verify the resolution for each image to avoid the ``pseudo level-20 images'' which are actually resized from level-19 images.
Considering the practical requirements of users, we also additionally collect level-19 images.
To be precise, \textbf{\ourdataset19} stands for the level-19 part, and \textbf{\ourdataset20} stands for the level-20 part.

\paragraph{Geographic Diversity}
The roads in different geographic locations exhibit significant differences in the surrounding natural environment, construction regulations, spatial distribution, road conditions, and appearance.
To ensure such diversities, we sample the geographic locations considering \textit{(i)} famous and typical cities around the world;
\textit{(ii)} terrain with different appearances such as plains, mountains, and deserts;
\textit{(iii)} regions with different road densities such as urban and suburban.
In this way, we ensure the overall diversity of our dataset by controlling the geographic diversity.
In the following, we take more factors into account to further improve the data quality and diversity.

\paragraph{Special Road Structures}
Unlike the common straight roads, some roads have special and complicated structures such as flyovers, roundabouts, and winding roads.
The lane instances of these regions demonstrate different shapes and distributions from the straight lanes.
To ensure the models trained on our data could well handle these structures, we manually search and collect some regions containing them.
Figure~\ref{fig:fig1} shows examples of special road structures.

\paragraph{Traffic Feature}
Here we use the term traffic feature to represent left-hand/right-hand traffic regulations and vehicle density.
The left-hand/right-hand regulations determine the direction of lanes.
The vehicle density has an impact on the difficulty of lane detection since vehicles may occlude the lanes.
During data collection, we deliberately consider the regions with different driving regulations and balance the samples with different vehicle densities.

\begin{figure}
    \centering
    \begin{subfigure}{0.463\textwidth}
      \centering 
      \includegraphics[height=4.8cm]{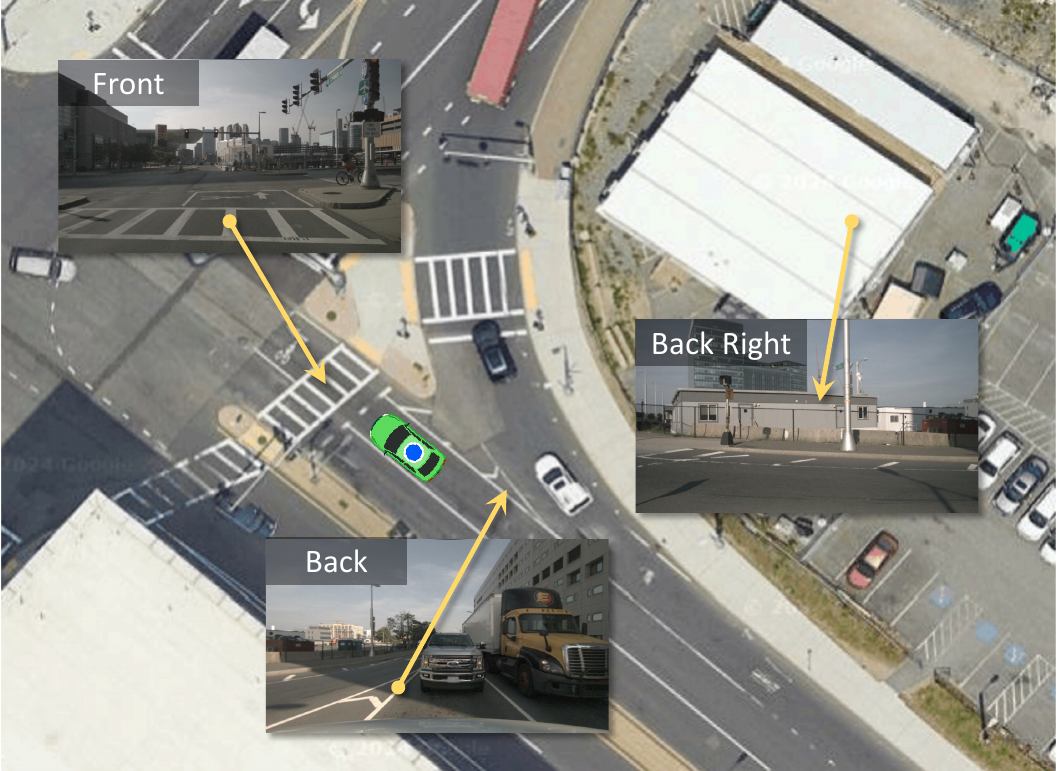}
        \caption{Landmark correspondences between a satellite image and a nuScenes image.}
        \label{fig:alignmentview}
    \end{subfigure}   %
 \hfill
     \begin{subfigure}{0.525\textwidth}
      \centering   
      \includegraphics[height=4.8cm]{pic/merged_image-boston-seapot-4096_trajectorynew14red_resized_0.5.pdf}
        \caption{Overlaying the driving trajectories from the nuScenes dataset onto \ourdataset{}~(Boston Seaport).
        }
        \label{fig:alignmentpose}
    \end{subfigure}
    \caption{\textbf{Alignment with driving benchmark}.}
           \label{fig:alignment}
\end{figure}

\paragraph{Alignment with Driving Benchmark}
To facilitate the map construction task in the field of autonomous driving, we make the collecting regions fully cover the driving region in famous driving benchmark nuScenes~\cite{caesar2020nuscenes} and Argoverse 2~\cite{chang2019argoverse}.
Since Argoverse 2 only provides the relative pose without GPS, we manually recognize a couple of landmarks of each city in Argoverse 2 and find the GPS coordinates of these landmarks.
Then the GPS coordinates of all samples in Argoverse 2 can be derived by the relative poses.
Figure~\ref{fig:alignment} showcases the alignment.

\subsection{Annotation} \label{sec:annotation}

After the data collection, we employ experienced annotators to carefully annotate the data at a fine-grained instance level.
The annotations are performed by a professional remote sensing imagery labeling team of approximately 50 annotators for labeling and 7 for quality checking.
The total cost of labeling is roughly \$72,000, \textit{i.e.}, \$19 per image. Note that, for labeling, the image size of \ourdataset{20} we collect is 4096 $\times$ 4096 and 2048 $\times$ 2048 for \ourdataset{19}.

\paragraph{Line representation.} To represent the lane lines with highly variable lengths and curvatures, we adopt vectorized polylines as the representation.
Each line contains a wealth of category and attribute information.
Meanwhile, the order of the points in the polylines defines the line direction.

\begin{figure}
    \centering
    \begin{subfigure}{0.413\textwidth}
      \centering   
      \includegraphics[height=4.7cm]{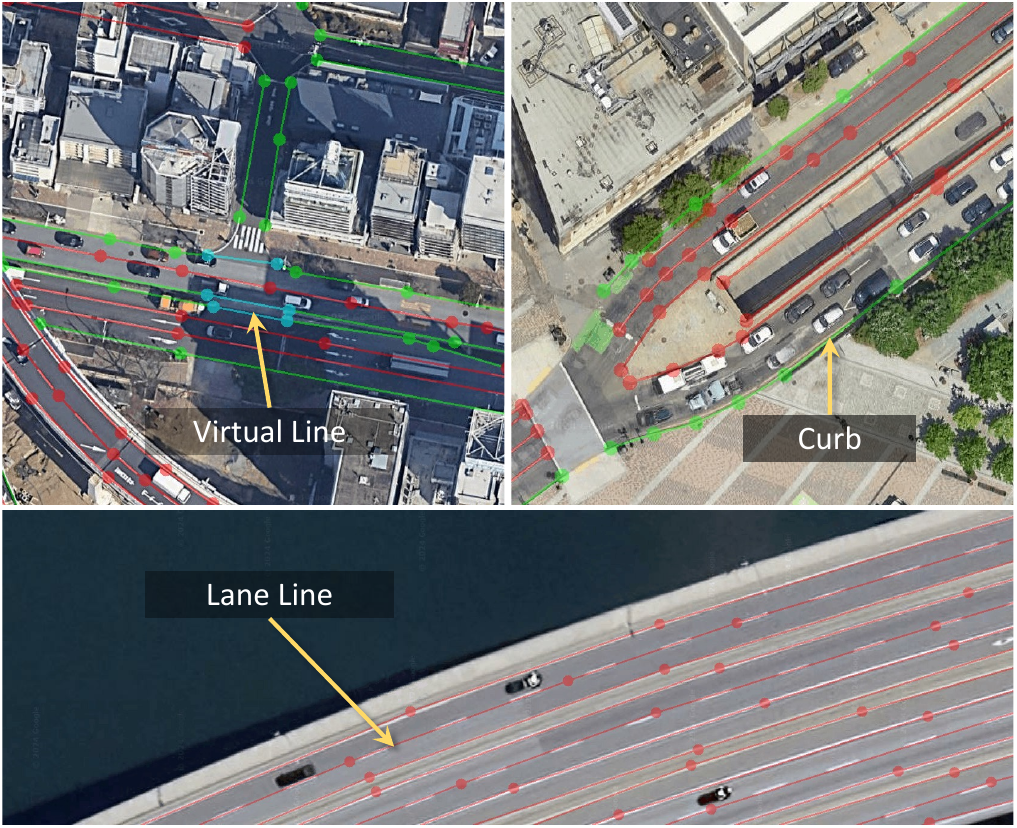}
        \caption{An example of three categories. }
        \label{fig:category}
    \end{subfigure}   %
\hfill
     \begin{subfigure}{0.58\textwidth}
      \centering   
      \includegraphics[height=4.7cm]{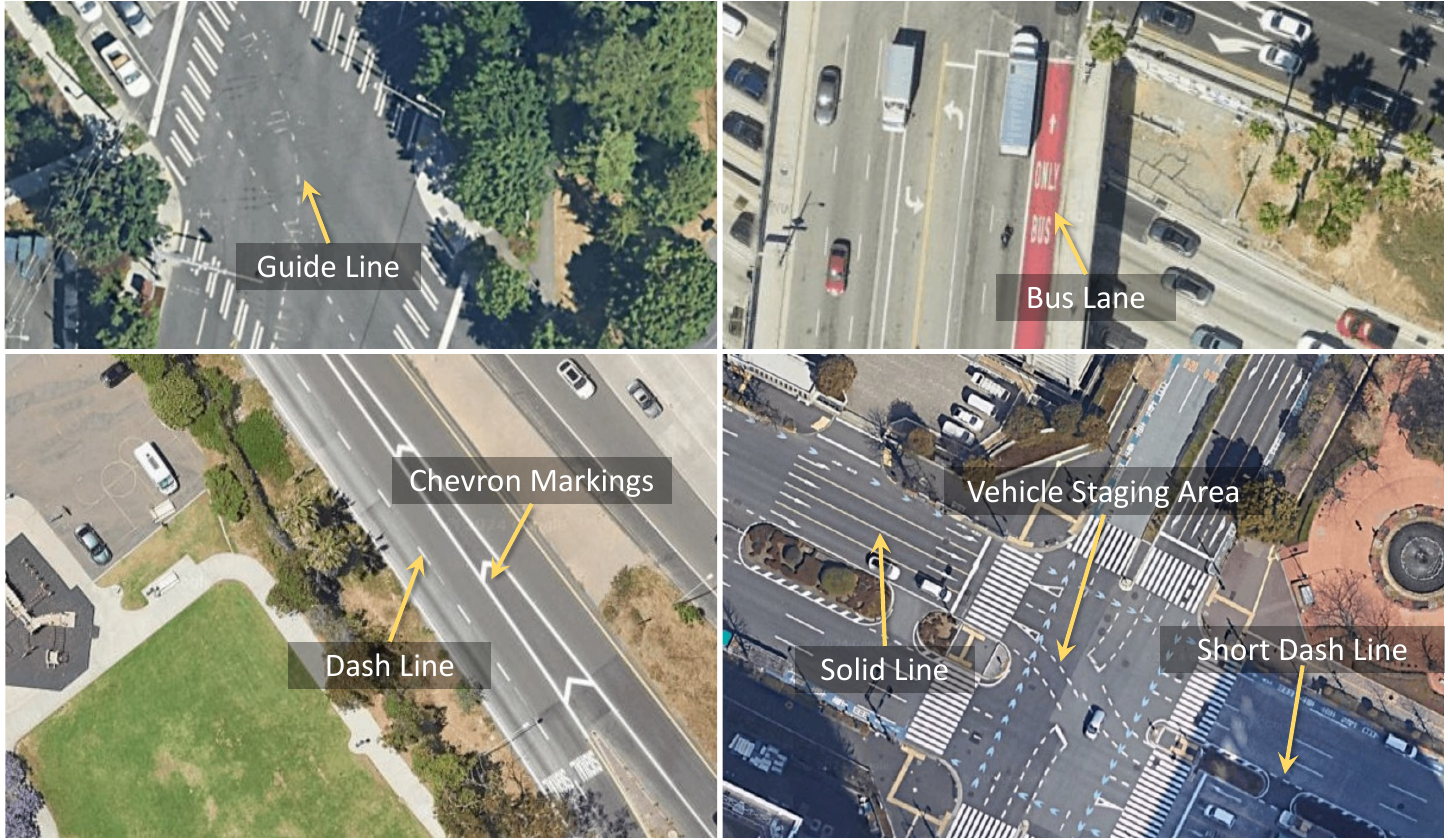}
        \caption{Examples of attributes.}
        \label{fig:attribute}
    \end{subfigure}
\caption{\textbf{Examples of categories and attributes}. 
}
\end{figure}

\paragraph{Line categories.} 
We first categorize all lines into three categories: \textbf{curb}, \textbf{lane line}, and \textbf{virtual line}.
A curb is the boundary of a road.
Lane lines are those visible lines forming the lanes.
A virtual line means that there is no lane line or curb here, but logically there should be a boundary to form a full lane.
For example, a fork in the road breaks a curb into two segments, then we use a virtual line to connect the two segments.
Figure~\ref{fig:category} shows examples of these three categories.

\paragraph{Where should it be labeled?}
Generally speaking, we label all three categories above.
However, there is usually occlusion in the satellite images.
If a line is partially occluded by buildings, trees, and vehicles but its complete shape can be inferred with high confidence, we annotate it. 
Fully occluded lines are not labeled. 
A special case is that overpasses usually occlude the bottom roads, where the occluded part is not labeled.
We show these cases in the supplementary materials.

\paragraph{Fine-grained line attributes.}
\begin{wrapfigure}{R}{0.5\textwidth}
\vspace{-13pt}
\begin{subfigure}{0.245\textwidth}
    \centering
    \includegraphics[height=3.5cm]{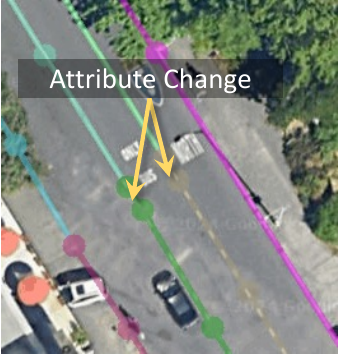}
    \caption{Attribute change.}
    \label{fig:sub1-instance-rule}
\end{subfigure}   %
\hfill
\begin{subfigure}{0.245\textwidth}
    \centering
    \includegraphics[height=3.5cm]{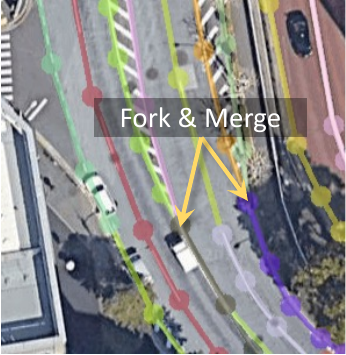}
    \caption{Lines fork and merge.}
    \label{fig:sub2-instance-rule}
\end{subfigure}
\caption{
\textbf{Definition of instances} (\textit{zoom in} for best viewing).
In (a), a change of line type results in two instances sharing the same point. 
In (b), a line should be divided into three instances when it is forked or merged.
}
\label{fig:rules}
\vspace{-20pt}
\end{wrapfigure}
Lines are assigned with a couple of attributes based on their appearance and functionalities.
Specifically, there are 8 attributes as follows:
\begin{itemize}[nosep, leftmargin=*]
    \item The colors of lines.
    \item The line type such as solid line, thick solid line, dashed line, and short dashed line.
    \item The number of lines such as single lines and double lines.
    \item Lines with special functionalities such as bus lanes, guide lines, lane-borrowing areas.
    \item Whether it is bi-directional.
    \item Whether it is the outermost boundary of the pavement.
    \item The level of occlusion.
    \item The level of clearness.
\end{itemize}
Figure~\ref{fig:attribute} shows some examples of different attributes.

\paragraph{Instances definition.} 

After defining the line attributes, we further define the instance using the following guidelines.
\textit{(i)} Different instances have different attributes. 
For example, if a solid line becomes a dashed line, we cut the line into two instances. Figure~\ref{fig:rules}\textcolor{red}{a} shows an example of such attribute change.
\textit{(ii)} When lines fork or merge (\textit{e.g.}, ``Y''-shape point), we break the lines into multiple instances as Figure~\ref{fig:rules}\textcolor{red}{b} shows.

\paragraph{Image-level tags.}
Furthermore, we provide \ourdataset20 with 4 additional image-level tags to describe general information, including image clearness, overall vehicle density, urban/suburban/rural, and the existence of special road structures. 
We present the details in the {supplementary materials}.

\subsection{Statistics of \ourdataset}
\label{sec:stats}
\begin{wrapfigure}{R}{0.5\textwidth}
\vspace{-15pt}
\begin{subfigure}{0.245\textwidth}
    \centering
    \includegraphics[height=2.65cm]{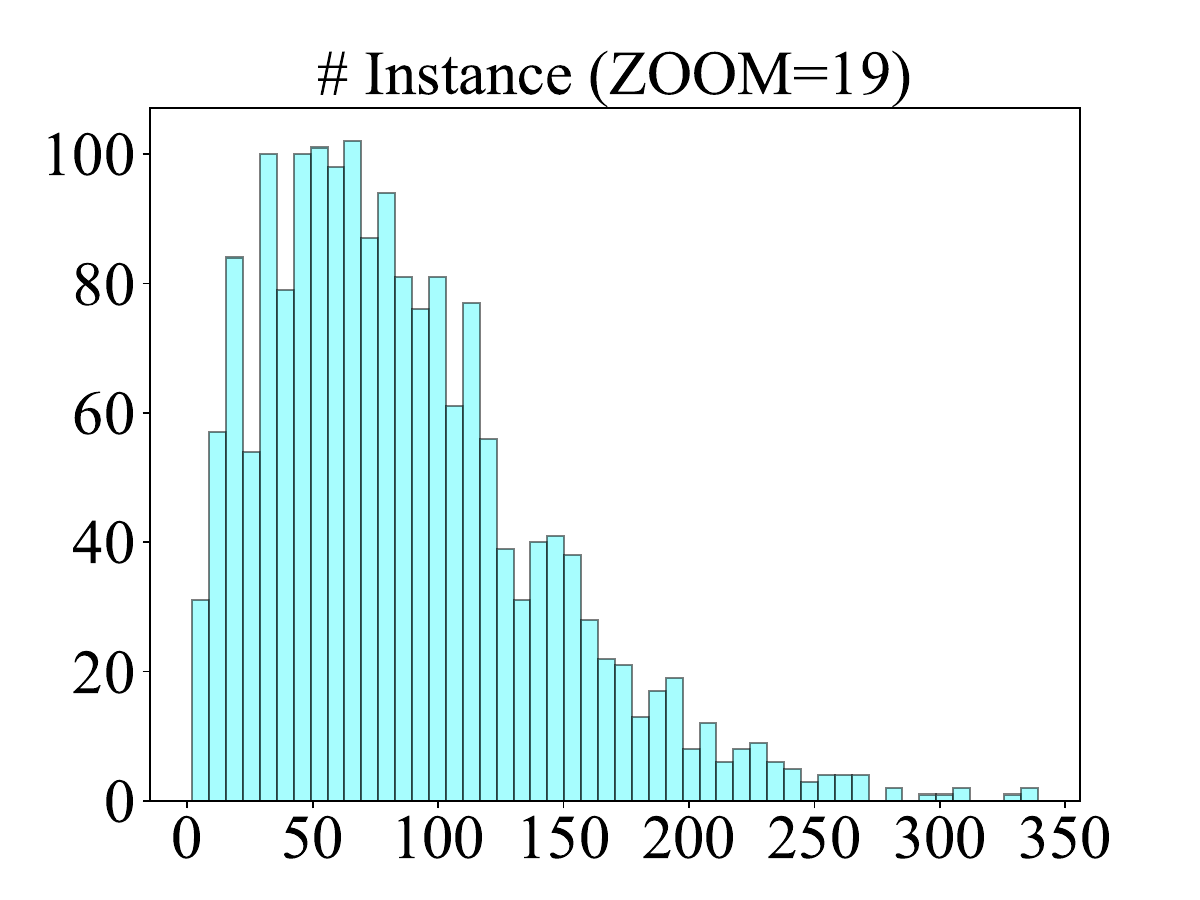}
    \label{fig:sub1-instance}
\end{subfigure}   %
\hfill
\begin{subfigure}{0.245\textwidth}
    \centering   
    \includegraphics[height=2.65cm]{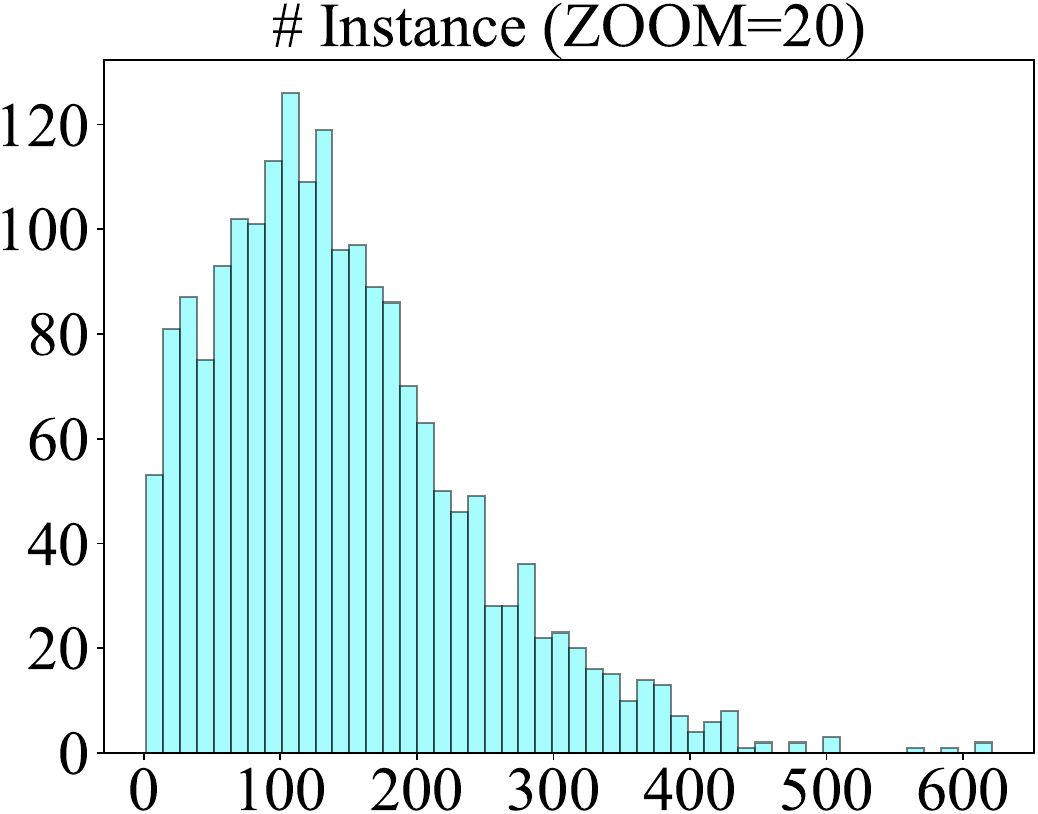}
    \label{fig:sub2-instance}
\end{subfigure}
\vspace{-20pt}
\caption{
Number of instances in each image in \ourdataset19 (left) and \ourdataset20 (right).
}
\label{fig:instance}
\end{wrapfigure}
\begin{figure}[t]
  \centering
    \begin{subfigure}{0.245\textwidth}
      \centering   
      \includegraphics[width=\linewidth]{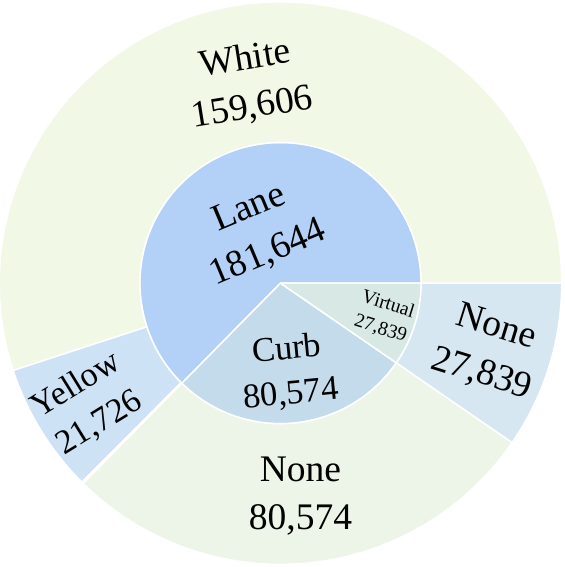}
        \caption{Color distribution.}
        \label{fig:subcolor-sun20}
    \end{subfigure}   %
    \hfill  
    \begin{subfigure}{0.245\textwidth}
      \centering   
      \includegraphics[width=\linewidth]{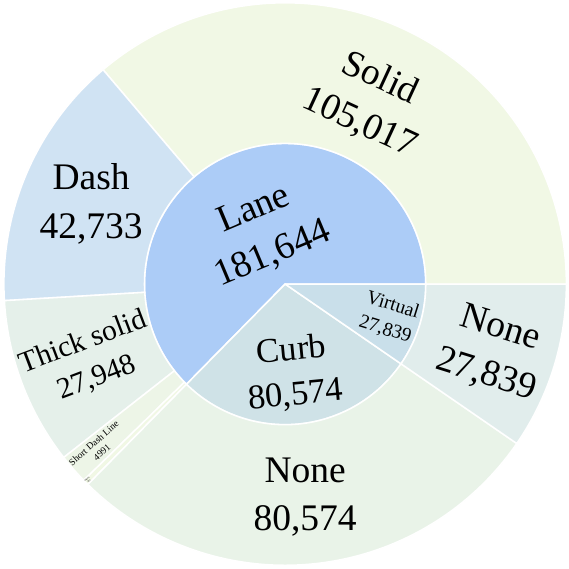}
        \caption{Line type distribution.}
        \label{fig:subtype-sun20}
    \end{subfigure}
        \begin{subfigure}{0.245\textwidth}
      \centering   
      \includegraphics[width=\linewidth]{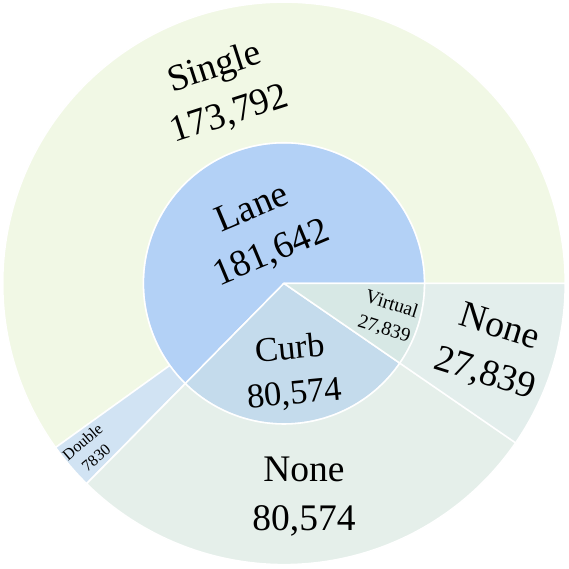}
        \caption{\# lines distribution.}
        \label{fig:subnum-sun20}
    \end{subfigure}
        \begin{subfigure}{0.245\textwidth}
      \centering   
      \includegraphics[width=\linewidth]{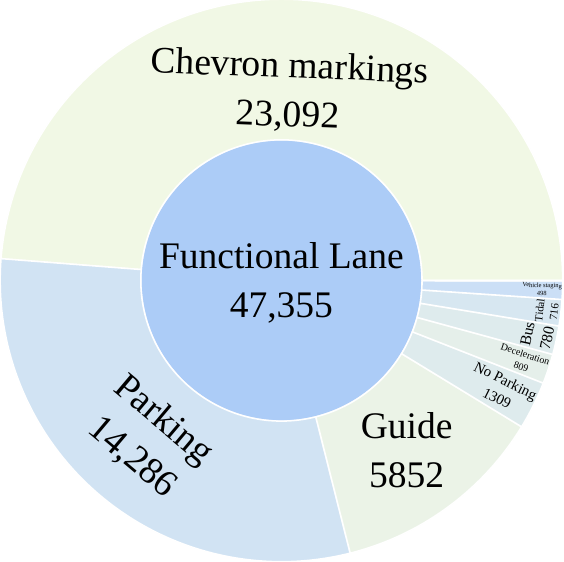}
        \caption{Function distribution.}
        \label{fig:subattr-sun20}
    \end{subfigure}
    \\
    \begin{subfigure}{0.245\textwidth}
      \centering   
      \includegraphics[width=\linewidth]{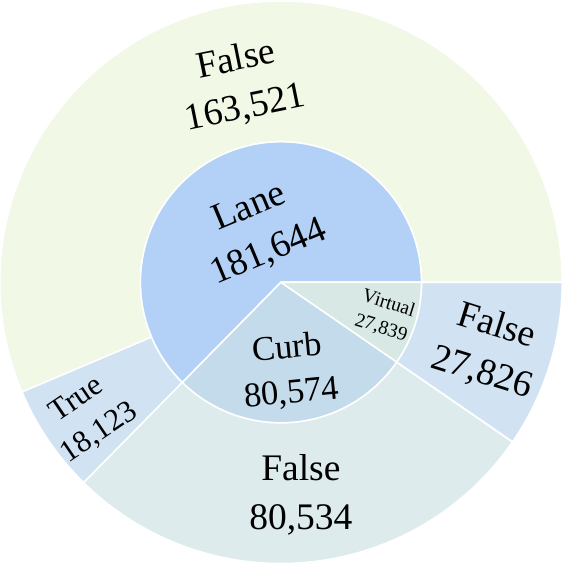}
        \caption{Bidirection distribution.}
        \label{fig:subbidir-sun20}
    \end{subfigure}   %
    \hfill  
    \begin{subfigure}{0.245\textwidth}
      \centering   
      \includegraphics[width=\linewidth]{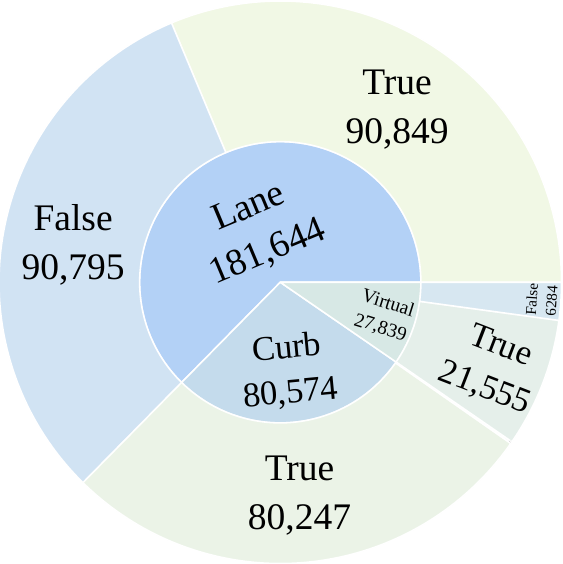}
        \caption{Boundary distribution.}
        \label{fig:subboundary-sun20}
    \end{subfigure}
        \begin{subfigure}{0.245\textwidth}
      \centering   
      \includegraphics[width=\linewidth]{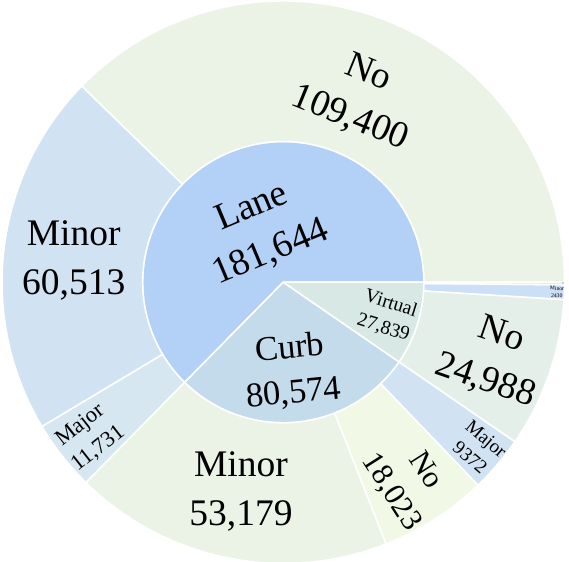}
        \caption{Occlusion distribution.}
        \label{fig:subocc-sun20}
    \end{subfigure}
        \begin{subfigure}{0.245\textwidth}
      \centering   
      \includegraphics[width=\linewidth]{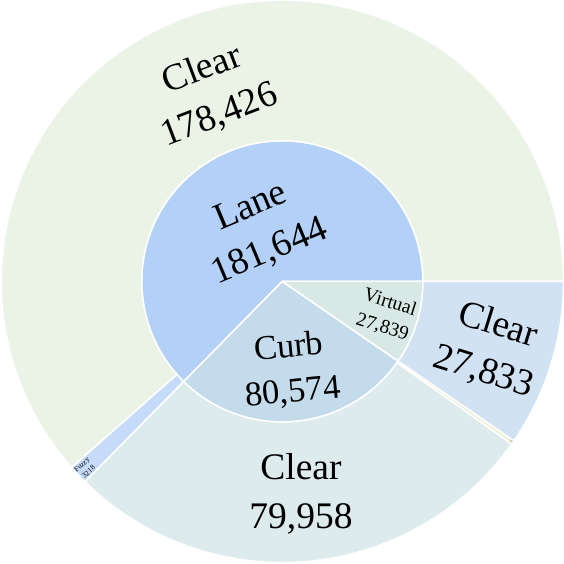}
        \caption{Clearness distribution.}
        \label{fig:subclear-sun20}
    \end{subfigure}
\caption{
Distribution of line attributes in  \ourdataset20. 
}
\vspace{-3mm}
\label{fig:sun}
\end{figure}
\begin{wrapfigure}{R}{0.5\textwidth}
\vspace{-12pt}
  \centering
    \begin{subfigure}{0.45\textwidth}
      \centering   
      \includegraphics[width=\linewidth]{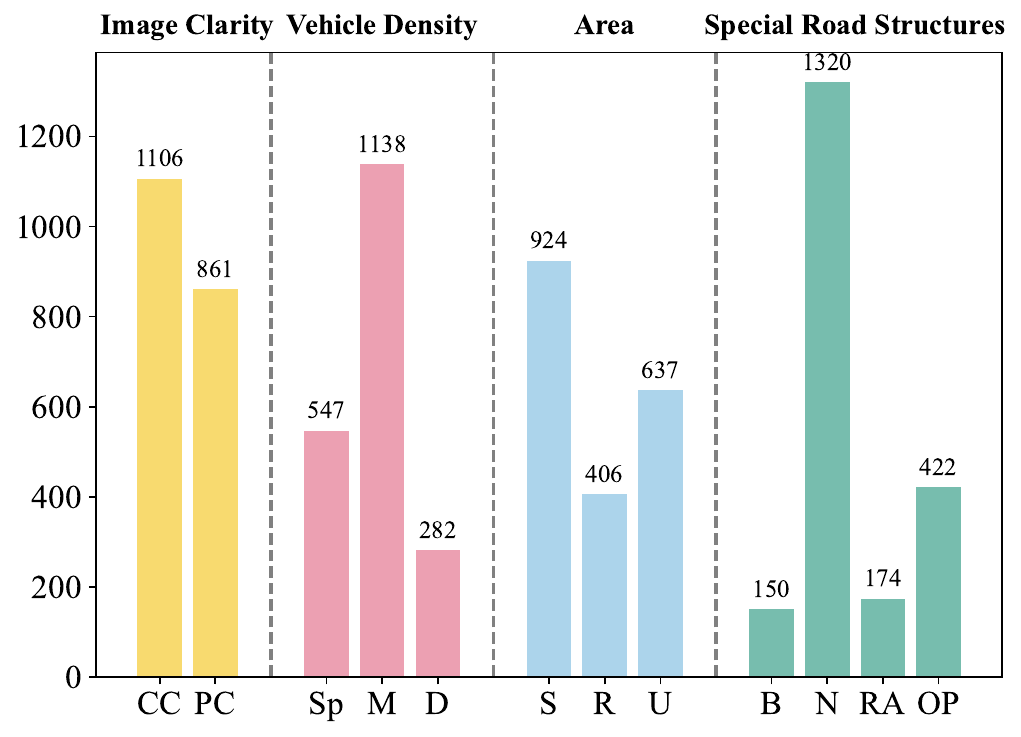}
        \label{fig:tag}
    \end{subfigure}   %

\vspace{-10pt}
\caption{
\textbf{Image-level tag distribution in \ourdataset20.}
CC = complete clear, PC = partially clear, Sp = sparse, M = moderate, D = dense, S = suburban, R = rural, U = urban, B = bridge, N = None, RA = roundabout, OP = overpass.
}
\label{fig:tag-dis}
\vspace{-10pt}
\end{wrapfigure}

In this section, we summarize the statistics of the established \ourdataset.
There are 1806 $2048\times2048$ images in \ourdataset19 and 1981 $4096\times4096$ images in \ourdataset20.
We randomly divided them into training set, validation set and testing set in the ratio of 6:2:2.
Figure~\ref{fig:instance} shows the number of instances in each image in \ourdataset19 and \ourdataset20.
The number of instances of most images is under 300.
Figure~\ref{fig:sun} illustrates the distributions of attributes. Most attributes have a non-uniform distribution, which reflects the real condition of the roads.
Figure~\ref{fig:tag-dis} shows the tag distribution in \ourdataset20.

\section{Instance-level Line Detection}

\subsection{Formulations and Baseline Method}

Given an input image $\mathbf{I} \in \mathbb{R}^{H \times W \times 3}$, where $H\times W$ indicates the input resolution, we aim to detect each line instance in the input image.
For each instance, we use polylines as the \textit{vectorized} representation and pixel-wise instance-level masks as the \textit{rasterized} representation.
A polyline is defined as a set of points $\mathbf{p} \in \mathbb{R}^{N \times 2}$, where $N$ means the number of sampled points.
$\mathbf{p}[:, 0]$ and $\mathbf{p}[:, 1]$ represent the $x$ and $y$ coordinates, respectively.
One can rasterize polylines to pixel-level masks by simply connecting adjacent points with lines with a pre-defined line width.

\begin{figure}[t]
    \centering
    \includegraphics[width=1\linewidth]{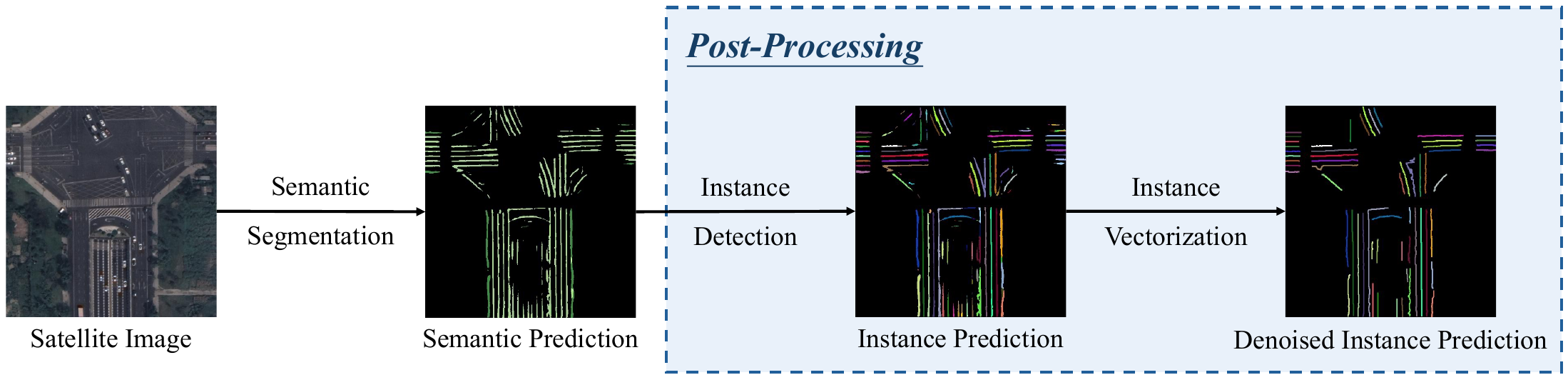}
    \caption{
    Illustration of our baseline method.
    }
    \vspace{-3mm}
    \label{fig:method}
\end{figure}

The instance-level line detection task needs to convert an RGB image $\mathbf{I}$ into a set of polylines $\{\mathbf{p}_i\}_{i=1}^{M}$, where $M$ is the number of instances.
We develop a simple baseline without whistle and bells illustrated in Figure~\ref{fig:method}.
We decompose this task into 3 steps: (1) semantic segmentation, (2) instance detection, and (3) instance vectorization.
(2) and (3) are post-processing techniques.
Detailed formulations for each step are provided as follows and implementation details are provided in the \textit{Supplementary materials}.

\textbf{Semantic segmentation} $f: \mathbb{R}^{H \times W \times 3} \to \{0,1,\cdots, C-1\}^{H\times W}$ aims to classify each pixel into a unique semantic category, where $C$ is the number of categories.
We denote $\mathbf{y} = f(\mathbf{I})$ as the segmentation result.
We adopt SegNeXt~\cite{guo2022segnext} as the segmentation network since it is quite efficient for high-resolution images.

\textbf{Instance detection} $g: \{0,1\}^{H\times W} \to \{0,1,\cdots, M\}^{H\times W}$ converts the binary segmentation mask of each category into a pixel-level mask, where each instance shares the same value within this mask.
Technically, we first extract the binary mask $\mathbf{y}_c \in \{0,1\}^{H \times W}$ of each category $c$ and then apply the watershed algorithm~\cite{watershed} to obtain instance-level masks.

\textbf{Instance vectorization} $h: \{0,1\}^{H \times W} \to \mathbb{R}^{N \times 2}$ aims to build a \textit{vectorized} representation of each instance, which contains (1) instance denoising, and (2) point sampling.
Instance denoising follows a simple sample-then-reconstruct pipeline.
Specifically, for each instance $i$ that belongs to category $c$, we first extract its binary mask $\mathbf{y}_c^i \in \{0, 1\}^{H \times W}$.
We subsequently sample $N$ points and obtain their coordinates $\mathbf{p}_c^i \in \mathbb{R}^{N \times 2}$.
Then, we simply connect adjacent points with lines and remove instances with fewer than 100 pixels, obtaining the denoised instance map $\hat{\mathbf{y}}_c^i = h(\mathbf{y}_c^i)$ for each instance $i$ that belongs to category $c$.
Note that $\hat{\mathbf{y}}_c^i$ is the final \textit{rasterized} representation of a line instance.
As for point sampling, we sample $N$ points from $\hat{\mathbf{y}}_c^i$, resulting in $\hat{\mathbf{p}}_c^i = \phi(\hat{\mathbf{y}}_c^i)$ to build the \textit{vectorized} representation.

\subsection{Evaluation}

\textbf{Semantic-level evaluation.}
We adopt the mean intersection over union (mIoU)~\cite{voc} over different categories as the metric to evaluate the performance at the semantic level.

\textbf{Instance-level evaluation.}
We adopt the average precision (AP) as the metric to evaluate the performance at the instance level.
Under instance segmentation settings, a common practice is to use the Mask AP (AP$^{\mathrm{M}}$) by leveraging a mask IoU threshold to identify whether an instance is a true positive sample or not~\cite{cheng2022masked, wang2023droppos, lin2014microsoft, carion2020end, he2017mask}.
However, line markings typically exhibit \textit{narrow} features and segmentation outputs frequently lack precision in edge definition.
Consequently, relying exclusively on Mask IoU thresholds may be excessively strict.
To this end, we incorporate Chamfer distance proposed by \citep{li2022hdmapnet} as an alternative. Chamfer distance between two sets of points $\mathbf{p} \in \mathbb{R}^{M\times 2}$ and $\mathbf{q} \in \mathbb{R}^{K \times 2}$ is defined as:
\begin{equation}
    D_{\mathrm{Chamfer}} (\mathbf{p}, \mathbf{q}) = \frac1M \sum_{i=1}^M \min_{j=1,2,\dots,K} d(p_i, q_j),
\end{equation}
where $d(\cdot, \cdot)$ is the Euclidean distance between two points.
Then, we get Chamfer AP (AP$^{\mathrm{C}}_D$) by choosing the distance threshold $D$. 

\begin{table}[t]
    \centering
    \caption{
    \textbf{Evaluation on \ourdataset{} validation set.}
    The line width is set to 6 pixels in \ourdataset20 and 3 pixels in \ourdataset19 by default.
    AP$^{\mathrm{M}}$ means that the mask IoU is used when determining true positives, while AP$^{\mathrm{C}}_D$ means Chamfer AP with a threshold of $D$ meters.
    AP$_{x}$ denotes that the threshold is set to $x$.
    AP$_{x:y}$ indicates the \textit{averaged} values, varying the threshold from $x$ to $y$.
    }
    \label{tab:exp_main}
    \vspace{2mm}
    \setlength{\tabcolsep}{1.8pt}
    \begin{tabular}{l|cccc|cccc}
    \toprule
    Dataset & AP$^{\mathrm{C}}_{\mathrm{0.9}}$ & AP$^{\mathrm{C}}_{\mathrm{1.5}}$ & AP$^{\mathrm{C}}_{\mathrm{3.0}}$ & AP$^{\mathrm{C}}_{\mathrm{4.5}}$ & AP$^{\mathrm{M}}_{\mathrm{50:95}}$ & AP$^{\mathrm{M}}_{\mathrm{50}}$ & AP$^{\mathrm{M}}_{\mathrm{75}}$ & mIoU \\
    \midrule
    \ourdataset19 & 16.04\tiny$\pm$0.35 & 22.68\tiny$\pm$0.35 & 26.88\tiny$\pm$0.52 & 29.18\tiny$\pm$0.22 & 3.66\tiny$\pm$0.15 & 10.66\tiny$\pm$0.44 & 1.45\tiny$\pm$0.12 & 28.71\tiny$\pm$0.38 \\
    \ourdataset20 & 20.30\tiny$\pm$0.21 & 25.93\tiny$\pm$0.35 & 29.50\tiny$\pm$0.40 & 31.38\tiny$\pm$0.43 & 6.98\tiny$\pm$0.21 & 16.05\tiny$\pm$0.32 & 5.26\tiny$\pm$0.13 & 33.69\tiny$\pm$0.45 \\
    \bottomrule
    \end{tabular}
\end{table}

\subsection{Main results}

Empirical results in Table~\ref{tab:exp_main} indicate that \textit{instance-level} line detection is a much more challenging task compared with semantic-level segmentation, since values of AP$^{\mathrm{C}}$ and AP$^{\mathrm{M}}$ are much lower than that of mIoU.

Fundamentally, this issue arises because instances are associated with fine-grained attributes while semantic segmentation only involves several categories.
Consequently, identifying true positive instance predictions becomes a notably difficult task.

\begin{figure}[t]
    \centering
    \includegraphics[width=1\linewidth]{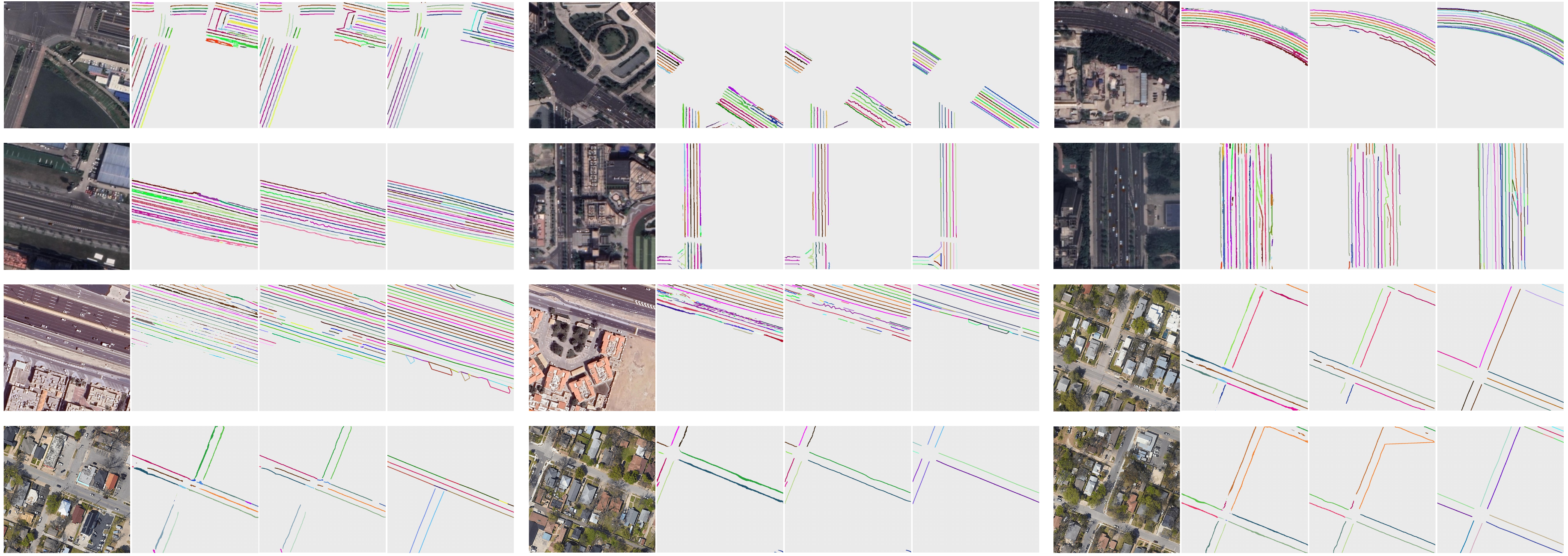}
    \caption{
    \textbf{Qualitative results} on \ourdataset19 (the first two rows) and \ourdataset20 (the last two rows) test split. 
    %
    %
    For each scene, we put \textbf{(a)} the input image, \textbf{(b)} the instance prediction, \textbf{(c)} the \textit{denoised} instance prediction, and \textbf{(d)} the annotation, \textbf{from left to right}, respectively.
    }
    \vspace{-3mm}
    \label{fig:visual}
\end{figure}

\textbf{Qualitative results.}
We provide qualitative results in Figure~\ref{fig:visual}.
Our method manages to detect line instances.
Although the visual results of our model are almost satisfactorily presented, they are accompanied by disappointingly low quantitative scores of AP$^{\mathrm{C}}$ and AP$^{\mathrm{M}}$.
Usually, our method fails to detect accurate line instances, when the segmentation performs poorly, such as failing to predict separate masks when attributes change, and adhered segmentation masks of two separate lines.
This disparity indicates that this benchmark is of substantial difficulty, necessitating deeper exploration of an effective end-to-end method.

\section{Potential Application in Autonomous Driving}
\vspace{-3mm}
To facilitate the map construction task in autonomous driving, we make \ourdataset{} deliberately cover the regions of nuScenes \cite{caesar2020nuscenes} and Argoverse 2 \cite{chang2019argoverse} datasets.
Recently, a line of work has proved the effectiveness and success of using additional map information to boost the performance of HDMap construction for autonomous driving.
MapEX \cite{sun2023mind} uses historically stored map information to optimize the construction of current local high-precision maps.
NMP \cite{xiong2023neuralmapprior} uses a neural map prior to boost the performance of VectorMapNet \cite{liu2022vectormapnet}.
\par
Some more closed work uses satellite image information to enhance map construction. 
SatforHDMap \cite{gao2024complementing} leverages satellite images as an additional input modality for MapTR~\cite{liao2022maptr}.
P-MapNet \cite{jiang2024p} extracts weakly aligned SDMap from OpenStreetMap \cite{OpenStre43:online}, and encode it as an additional feature for MapTR, achieving better performance.
Our \ourdataset{} has the potential to advance this field.
We take SatforHDMap as an example. We use intersection-over-union (IoU) as the metric following semantic map learning \cite{li2022hdmapnet}.
Let $\mathcal{D}_1,\mathcal{D}_2 \in \mathbb{R}^{H\times W\times D}$ be dense predictions of shapes, $H$ and $W$ are the height and width of the grid, $D$ is the number of categories.
IoU can be expressed as follows:

\begin{equation}
\text{IoU}\left(\mathcal{D}_1, \mathcal{D}_2\right)=\frac{\left|\mathcal{D}_1 \cap \mathcal{D}_2\right|}{\left|\mathcal{D}_1 \cup \mathcal{D}_2\right|},
\end{equation}
where $|\cdot|$ is the size of the set.
Table~\ref{tab:hdmap} shows the results using SatforHDMap\cite{gao2024complementing} with \ourdataset.
The use of corresponding satellite imagery significantly enhanced the performance of online map construction, demonstrating the potential superiority of this approach.

\begin{table}[h]
    \centering
    \setlength{\tabcolsep}{10pt}
    \caption{Evaluation of semantic map segmentation on nuScenes validation set.}
    \begin{tabular}{l|cccc}
    \toprule
    Method & Divider & Crossing & Boundary & All \\
    \midrule
     HDMapNet \cite{li2022hdmapnet}    & 40.6& 18.7& 39.5& 32.9 \\
    SatforHDMap \cite{gao2024complementing} &\textbf{50.2}  & \textbf{53.2} & \textbf{49.4}&  \textbf{50.9}\\
     \bottomrule
    \end{tabular}
    \label{tab:hdmap}
\end{table}

\section{Availability and Maintenance}\label{sec:availability}
\vspace{-3mm}
\ourdataset{} is collected from Google Maps, which is publicly available. 
According to the regulations of Google, the use of this dataset is restricted to \textbf{non-commercial research}.
Our project page is \url{https://opensatmap.github.io}, which contains the full dataset with annotations and code repository.
The repository contains the full-stack tools to use this dataset, including code for collecting images from Google Maps, documents, baseline models, and evaluation tools.
We encourage the community to further explore more tracks and effective methods using \ourdataset{}.
This dataset will be maintained over the long term and continually iterated upon.

\section{Limitations and Potential Social Impact} \label{sec:limitations}
\vspace{-3mm}
\ourdataset{} has the following limitations.
\textit{(i)} \ourdataset~ is collected from Google Maps, which is not updated in real time, and certain areas may be outdated and not reflect the current conditions.
Besides, the timing of captured remote sensing images may not be aligned with nuScenes and Argoverse dataset. 
Although we noticed this potential misalignment on road structures and asked the annotators to check the consistency of road structures between our data and the data in the driving datasets during annotating process, there are still very few inconsistencies.
\textit{(ii)} High-resolution (level-20) images in certain areas are not available or may be covered by the cloud, limiting the diversity of collection locations to some extent.
\textit{(iii)} Although the annotators are quite professional and we conduct strict sanity checks, the subjectivity of annotators may inevitably lead to slightly inconsistent annotation results since this project employs around 60 annotators. 
In addition, the inconsistent annotation can also stem from the different training and expertise levels of the annotators.
A potential social impact is that it might cause a safety risk for driving if the model is directly trained on our dataset, which contains outdated data as discussed above.

\section{Conclusion}
\vspace{-3mm}
In summary, we introduce \ourdataset{}, a large-scale, high-resolution, geographically diverse satellite dataset with detailed annotations and alignment with driving benchmarks. 
To validate the utility of \ourdataset{}, we construct the instance-level line detection track and provide a baseline for it.
Moreover, we use \ourdataset{} to improve the performance of online map construction, which shows the great potential of autonomous driving.
We believe that our carefully annotated high-quality \ourdataset{} can serve as the foundation for various applications, including lane detection, city-scale map construction, and autonomous driving.

\begin{ack}
This work was supported in part by the National Key R\&D Program of China (No. 2022ZD0116500), the National Natural Science Foundation of China (No. U21B2042, No. 62320106010), and in part by the 2035 Innovation Program of CAS, and the InnoHK program, and in part by the Tencent Maps Collaborative Research Project.
\end{ack}

\section*{Checklist}
\begin{enumerate}

\item For all authors...
\begin{enumerate}
  \item Do the main claims made in the abstract and introduction accurately reflect the paper's contributions and scope?
    \answerYes{}
  \item Did you describe the limitations of your work?
    \answerYes{See in \Cref{sec:limitations}.}
  \item Did you discuss any potential negative societal impacts of your work?
    \answerYes{See in \Cref{sec:limitations}}
  \item Have you read the ethics review guidelines and ensured that your paper conforms to them?
    \answerYes{}
\end{enumerate}

\item If you are including theoretical results...
\begin{enumerate}
  \item Did you state the full set of assumptions of all theoretical results?
    \answerNA{We are not including these.}
	\item Did you include complete proofs of all theoretical results?
    \answerNA{We are not including these.}
\end{enumerate}

\item If you ran experiments (e.g. for benchmarks)...
\begin{enumerate}
  \item Did you include the code, data, and instructions needed to reproduce the main experimental results (either in the supplemental material or as a URL)?
    \answerYes{See our project page.}
  \item Did you specify all the training details (e.g., data splits, hyperparameters, how they were chosen)?
    \answerYes{See the supplemental material.}
	\item Did you report error bars (e.g., with respect to the random seed after running experiments multiple times)?
    \answerYes{See in Table~\ref{tab:exp_main}}
	\item Did you include the total amount of compute and the type of resources used (e.g., type of GPUs, internal cluster, or cloud provider)?
    \answerYes{See the supplemental material.}
\end{enumerate}

\item If you are using existing assets (e.g., code, data, models) or curating/releasing new assets...
\begin{enumerate}
  \item If your work uses existing assets, did you cite the creators?
    \answerYes{}
  \item Did you mention the license of the assets?
    \answerYes{}
  \item Did you include any new assets either in the supplemental material or as a URL?
   \answerYes{See the project page.}
  \item Did you discuss whether and how consent was obtained from people whose data you're using/curating?
    \answerYes{See \Cref{sec:availability}.}
  \item Did you discuss whether the data you are using/curating contains personally identifiable information or offensive content?
    \answerYes{Sec.~\ref{sec:collection} and Sec.~\ref{sec:availability}.}
\end{enumerate}

\item If you used crowdsourcing or conducted research with human subjects...
\begin{enumerate}
  \item Did you include the full text of instructions given to participants and screenshots, if applicable?
    \answerYes{See \Cref{sec:annotation}.}
  \item Did you describe any potential participant risks, with links to Institutional Review Board (IRB) approvals, if applicable?
    \answerNA{This work does not use human subjects.}
  \item Did you include the estimated hourly wage paid to participants and the total amount spent on participant compensation?
    \answerYes{See \Cref{sec:annotation}.}
\end{enumerate}

\end{enumerate}
\clearpage
\appendix

In this Supplementary Material, we:
\begin{itemize}
    \item Provide a detailed description of the dataset following guides on dataset publication \cite{gebru2021datasheets}.
    \item Provide the details and results on instance-level line detection and satellite-enhanced online map construction for autonomous driving. 
\end{itemize}

\section{\ourdataset{} Description}
\subsection{Datasheets}

\subsubsection{Motivation}
\begin{enumerate}
    \item \textbf{For what purpose was the dataset created?} Was there a specific task in mind? Was there a
specific gap that needed to be filled? Please provide a description.\\
\ourdataset{} is created to parse fine-grained road structures from satellite images and can serve as a foundation for various applications, including city-scale map construction, lane line detection, and autonomous driving, \textit{etc.}.
Existing benchmarks \cite{liu2018roadnet, bastani2018roadtracer, demir2018deepglobe,cheng2017automatic} have made attempts to handle this task.
However, their limitations including coarse annotations, low resolution, small scale and unalignment with autonomous driving benchmarks prevent them from effectively supporting large-scale and fine-grained map construction.
 \item \textbf{Who created the dataset and on behalf of which entity?}\\
The dataset was developed by a consortium of  ML researchers from CASIA, HKISI and employees from Tencent Maps listed in the author list. 

 \item \textbf{Who funded the creation of the dataset?}\\
 This work was supported in part by the National Key R\&D Program of China (No. 2022ZD0116500), the National Natural Science Foundation of China (No. U21B2042, No. 62320106010), and in part by the 2035 Innovation Program of CAS, and the InnoHK program, and in part by the Tencent Maps Collaborative Research Project.
\end{enumerate}

\subsubsection{Composition}
\begin{enumerate}
    \item \textbf{What do the instances that comprise the dataset represent?}\\
Our dataset contains satellite images and annotations.
For each image, we annotate lane lines with three categories and eight attributes.

\item \textbf{How many instances are there in total (of each type, if appropriate)?}\\
There are 1806 images in \ourdataset{19} with image size 2048 $\times$ 2048 and 1981 images in \ourdataset{20} with image size 4096 $\times$ 4096.
For each image, we use polylines to annotate lane lines and 8 attributes (color, line type, number of lines, line function, bi-direction, boundary, occlusion and clearness) for each line.
There are 446,645 lines in total.

\item \textbf{Does the dataset contain all possible instances or is it a sample (not necessarily random) of instances from a larger set?}\\
Yes, it contains all possible instances.
\item \textbf{Is there a label or target associated with each instance?}\\
Yes.
\item \textbf{Is any information missing from individual instances?}\\
Yes. In \ourdataset{19}, we can not provide the GPS information because of the regulation in China. However, this does not affect the use of our dataset.
\item \textbf{Are relationships between individual instances made explicit?}\\
Yes. Our dataset is well-organized and the relationships between instances are explicit.
\item \textbf{Are there recommended data splits (e.g., training, development/validation, testing)?}\\
Yes. We have already done this for users when the dataset releases.
\item \textbf{Are there any errors, sources of noise, or redundancies in the dataset?} \\
Yes. Noise comes from some poor-quality images from Google Maps.
\item \textbf{Is the dataset self-contained, or does it link to or otherwise rely on external resources (e.g., websites, tweets, other datasets)?}\\
The dataset is self-contained.
\item \textbf{Does the dataset contain data that might be considered confidential?}\\
No.
\item \textbf{Does the dataset contain data that, if viewed directly, might be offensive, insulting, threatening, or might otherwise cause anxiety?}\\
No.    
\end{enumerate}

\subsubsection{Collection Process}
\begin{enumerate}
    \item \textbf{How was the data associated with each instance acquired?} Was the data directly
observable (e.g., raw text, movie ratings), reported by subjects (e.g., survey responses), or
indirectly inferred/derived from other data (e.g., part-of-speech tags, model-based guesses for
age or language)? If the data was reported by subjects or indirectly inferred/derived from other data, was the data validated/verified?\\
The data is acquired from Google Maps using static public API. The collection process is in accordance with Google Maps terms. Please refer to \url{https://maps.google.com/help/terms_maps/} for more details.
    \item \textbf{What mechanisms or procedures were used to collect the data
(e.g., hardware apparatuses or sensors, manual human curation,
software programs, software APIs)?}\\
We use one NVIDIA A30 to run a program with the Maps Static API in Google Maps.
We will provide the source code in our GitHub repository.
The collection process is in accordance with Google Maps terms. Please refer to \url{https://maps.google.com/help/terms_maps/} for more details.
    \item \textbf{Who was involved in the data collection process (e.g., students, crowdworkers, contractors) and how were they compensated?}\\
    The authors in the author list collect the data and a professional remote sensing imagery labeling team of approximately 50 annotators label the data.
    The total cost of labeling is about \$72,000.
    For each crowdworker, the hourly rate is \$10.
    
    \item \textbf{Over what timeframe was the data collected?}\\
    The data was collected over 7 weeks, during the Spring of 2024 (March 15th, 2024 through May 10th, 2024).

    \item \textbf{Were any ethical review processes conducted (e.g., by an institutional review board)?}\\
    Yes. We obtain the publicly available images from Google Maps. Google has conducted the ethical review processes for the data.

\end{enumerate}
\subsubsection{Preprocessing/Labeling}
\begin{enumerate}
    \item \textbf{Was any preprocessing/cleaning/labeling of the data done?}\\
    Yes. We manually cleaned the images with poor quality (blurred, distorted, spliced, \textit{etc}. and without roads.).
    \item \textbf{Was the “raw” data saved in addition to the preprocessed/cleaned/labeled data (e.g., to support unanticipated future uses)?}\\
    Yes. We will provide it along with the release of the dataset.
    \item \textbf{Is the software that was used to preprocess/clean/label the data
available?}\\
The annotators use software to help the annotation process, and the software is developed themselves.

\end{enumerate}

\subsubsection{Uses}
\begin{enumerate}
    \item \textbf{Has the dataset been used for any tasks already?}\\
    No, this dataset has not been used for
any tasks yet.
\item \textbf{What (other) tasks could the dataset be used for?}\\
It could be used to conduct instance-level line detection, satellite-enhanced online map construction for autonomous Driving, super-resolution image reconstruction, \textit{etc.}.
\item \textbf{Are there tasks for which the dataset should not be used?}\\
No.
\item \textbf{Is there anything about the composition of the dataset or the way
it was collected and preprocessed/cleaned/labeled that might impact future uses?}\\
Yes. We only annotated the lane lines in the dataset, leaving a large amount of information unlabeled (\textit{e.g.}, buildings, 
cars \textit{etc.}), which limits the further uses of this dataset. 
We have already discussed this in Sec. \textcolor{red}{7}.
\end{enumerate}
\subsubsection{Distribution}
\begin{enumerate}
    \item \textbf{Will the dataset be distributed to third parties outside of the entity (e.g., company, institution, organization) on behalf of which
the dataset was created?}\\
Yes, the dataset is open to the public.
\item \textbf{How will the dataset be distributed (e.g., tarball on website,
API, GitHub)?}\\
We intend to distribute this dataset through Hugging Face and tarball on website.
The code for baselines will be released on GitHub.
\item \textbf{When will the dataset be distributed?}\\
It will be distributed in fall 2024.
\item \textbf{Will the dataset be distributed under a copyright or other intellectual property (IP) license, and/or under applicable terms of use
(ToU)?}\\
The dataset will be licensed under a Creative Commons CC-BY-NC-SA 4.0 license.  
Meanwhile, the use of the images from Google Maps must respect the "Google Maps" terms of use (\url{https://about.google/brand-resource-center/products-and-services/geo-guidelines/}, \url{https://maps.google.com/help/terms_maps/}).

\item \textbf{Have any third parties imposed IP-based or other restrictions on
the data associated with the instances?}\\
No.
\item \textbf{Do any export controls or other regulatory restrictions apply to
the dataset or to individual instances?}\\
No.
\end{enumerate}

\subsubsection{Maintenance}
\begin{enumerate}
    \item \textbf{Who will be supporting/hosting/maintaining the dataset?}\\
    All the authors will be supporting/hosting/maintaining the dataset in the long term.
    \item \textbf{How can the owner/curator/manager of the dataset be contacted (e.g., email address)?}\\
    Through our project page \url{https://opensatmap.github.io/} which contains our email addresses or GitHub issues.
    \item \textbf{Is there an erratum?}\\
    Not yet.
    \item \textbf{Will the dataset be updated?}\\
 Yes, the datasets will be updated whenever necessary to ensure accuracy, and announcements will be made accordingly.
    All the updates can be found in our project page \url{https://opensatmap.github.io/}.
    \item \textbf{If the dataset relates to people, are there applicable limits on the
retention of the data associated with the instances?}\\
Not applicable.
\item \textbf{Will older versions of the dataset continue to be supported/hosted/maintained?}\\
Yes, older versions of the dataset will continue to be supported/hosted/maintained.
\item \textbf{If others want to extend/augment/build on/contribute to the
dataset, is there a mechanism for them to do so?}\\
Yes. They can contact us from our project page or post a GitHub issue.
\end{enumerate}

\subsection{Dataset Annotation Documentation} \label{sec:doc}
In this subsection, we provide a detailed description of our instance-level and image-level annotation rules, which helps the dataset consumers to better understand and use our dataset.

\subsubsection{Overview}
\ourdataset{} is a high-resolution, geographically diverse, large-scale, satellite image dataset with fine-grained instance-level annotations.
Besides, in order to advance the map construction in autonomous driving, \ourdataset{} covers the regions of nuScenes \cite{caesar2020nuscenes} and Argoverse 2 \cite{chang2019argoverse} datasets.

\subsubsection{Instance-level Annotation Rules}
We use vectorized polylines to represent a line instance.
We first categorize all lines into three categories: \textbf{curb, lane line,} and \textbf{virtual line}.
A curb is the boundary of a road. Lane lines are those visible lines forming the lanes. A virtual line means no lane line or curb here, but logically there should be a boundary to form a full lane.

Each line is assigned a couple of attributes based on its appearance and functionalities. Specifically, there are 8 attributes as follows:
\begin{itemize}
    \item Color: White, Yellow, Others, None. 
    \item Line type: Solid line, Thick solid line, Dashed line, Short dashed line, Others, None. 
    \item Number of lines: Single line, Double line, Others, None.
    \item Function: Chevron markings, No parking, Deceleration lane, Bus lane, Others, Tidal lane, Parking spaces, Vehicle staging area, Guidance line,  Lane-borrowing area.
    \item Bidireciton: True, False.
    \item Boundary: True, False.
    \item Occlusion: No occlusion, Minor occlusion, Major occlusion.
    \item Clearness: Clear, Fuzzy.
\end{itemize}
Note that there is no man-made visible line on curbs and virtual lines, so we annotate their colors, line types, numbers of lines, and functions as None.
In terms of line colors, line types, numbers of lines, functions, boundaries, and clearness, they are apparent and easy to distinguish.

In the following, we give more details about bidirection and occlusion.
For bidirection, the order of the points in the polylines defines the line direction. 
We can easily determine the direction of the lines by traffic flow, arrows, \textit{etc}.
However, it is hard to tell the directions of some lines, \textit{e.g.}, the center line of a road, which is bidirectional. 
Therefore, we design bidirection attribute to handle this condition.
For occlusion, we use the ratio of the lines to be blocked by trees, cars, and buildings.
When the ratio is greater than 50\%, we give it major occlusion.
When the ratio is smaller than 50\%, we label minor occlusion.
When the line is fully visible, we label no occlusion.

After defining the line attributes, we give the definition of each instance using the following rules.
Although we have already talked about the definition of the instance in Sec.~\textcolor{red}{3.2}, here is a more detailed discussion for better understanding.
The detailed rules of each tag are shown below.
\textit{(i)} Different instances have different attributes. For example, if a solid line becomes a dashed line, we cut the line into two instances. Please refer to Figure~\ref{fig:instance}\textcolor{red}{a} for an example.
\textit{(ii)} When a line fork or merge (\textit{e.g.}, “Y”-shape point), we break it into three instances. Please refer to Figure~\ref{fig:instance}\textcolor{red}{b} for an example.

In addition to these generalized rules above, there are some rules for specific situations. 

\textit{(i)} Occlusion inference. If a line is partially occluded by buildings, trees, and vehicles but its complete shape can be inferred with high confidence, we will annotate it. 
Fully occluded lines are not labeled. However, the overpass is a special case and we will discuss it later.

\textit{(ii)} Overpass. When there is a multi-level interchange, the upper-level overpass will obscure the lane lines on the lower level.
Although in this case, the underlying lane lines can be inferred,
the portion covered by the upper level of the interchange is \textit{not labeled }to avoid conflicts with the upper ones.
Besides, the lower overpass breaks off at this point for two instances.

\subsubsection{Image-level Annotation Rules}
We provide \ourdataset{20} with 4 additional image-level tags including image clarity, vehicle density, settlement type and special road structure to describe general information of each image.
The following are detailed rules for each tag.

\begin{figure}
    \centering
    \includegraphics[width=\linewidth]{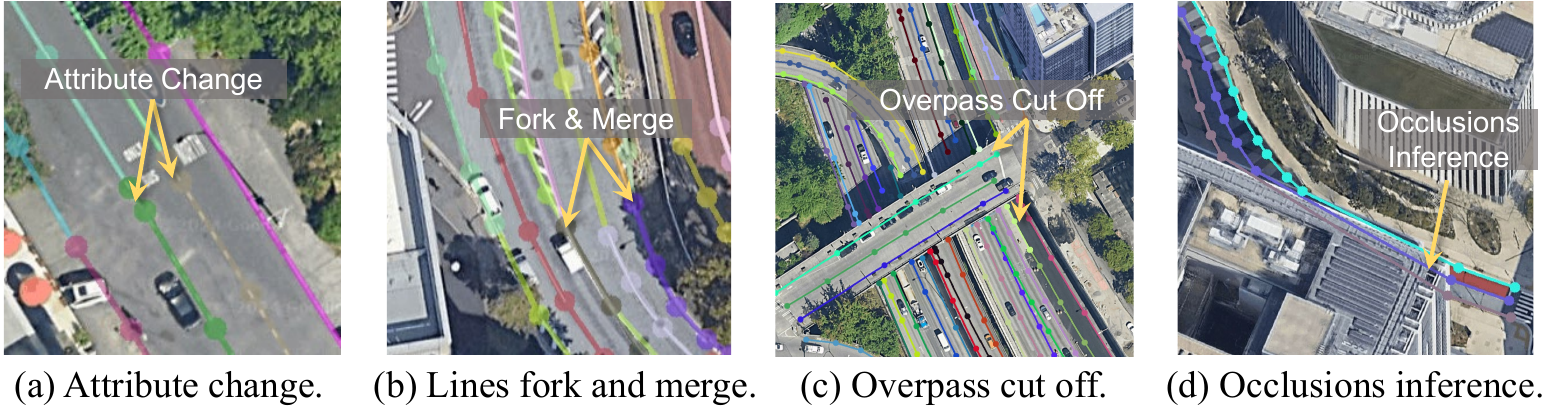}
    \caption{\textbf{Definition of instances} (\textit{zoom in} for best viewing).
In (a), a change of line type results in two instances sharing the same point. 
In (b), a line should be divided into three instances when it is forked or merged.
(c) shows an example of multi-level interchange. 
(d) shows the annotation inferences caused by the occlusion of the buildings.
}
\end{figure}

\textbf{Image clarity.} This attribute is used to describe the clarity of the image. 
\begin{itemize}
    \item Completely clear: Clear view of lane lines and surroundings.
    \item Partially clear: Some lane lines are clear, some are blurry or have ghosting, \textit{etc.} 
    Less than 30\% of the lane lines in the image are blurred.
    \item Not clear: More than 30\% of the roads are unclear. For images with this tag, we will not annotate them and remove them from \ourdataset{}.
\end{itemize}

\begin{figure}
    \centering
    \includegraphics[width=\linewidth]{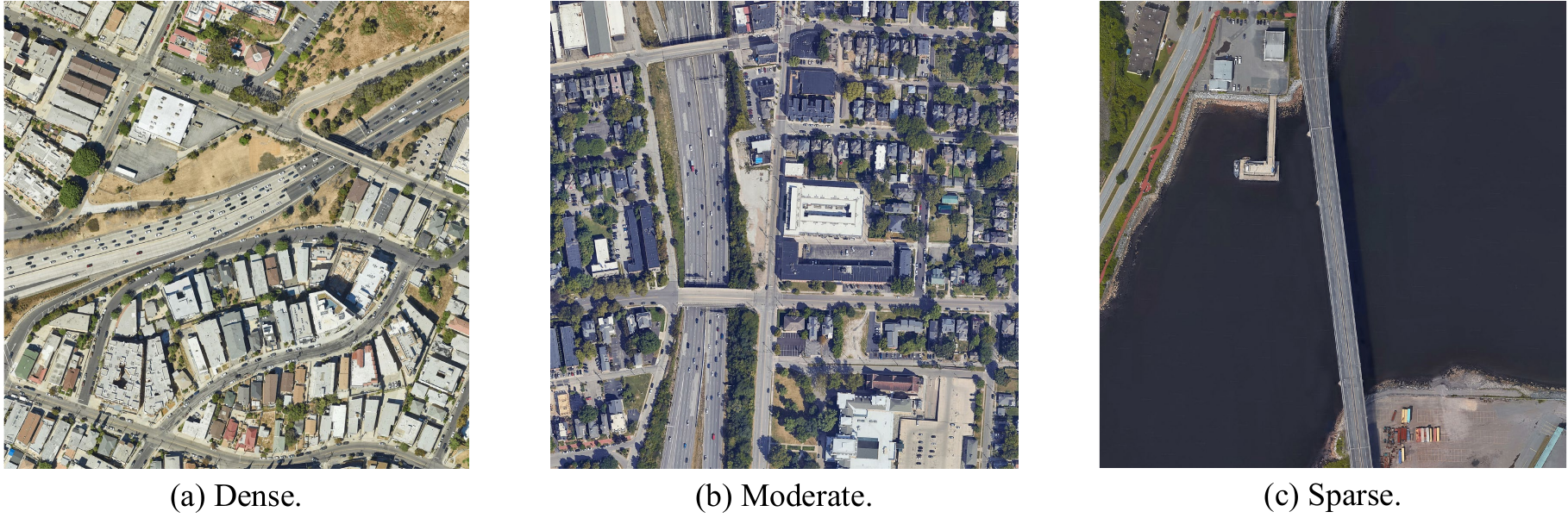}
    \caption{Vehicle density examples.}
    \label{fig:car}
\end{figure}

\textbf{Vehicle density.} We have three levels to describe the vehicle density of an image. Figure~\ref{fig:car} shows an example of each case.

\begin{itemize}
    \item Dense: The distance between cars is less than 5 vehicle lengths.
    \item Moderate: There are some cars on the road and the distance between cars is greater than 5 vehicle lengths.
    \item Sparse: Only less than 20\% of major roads have cars and the rest have no cars.
\end{itemize}

\begin{figure}
    \centering
    \includegraphics[width=\linewidth]{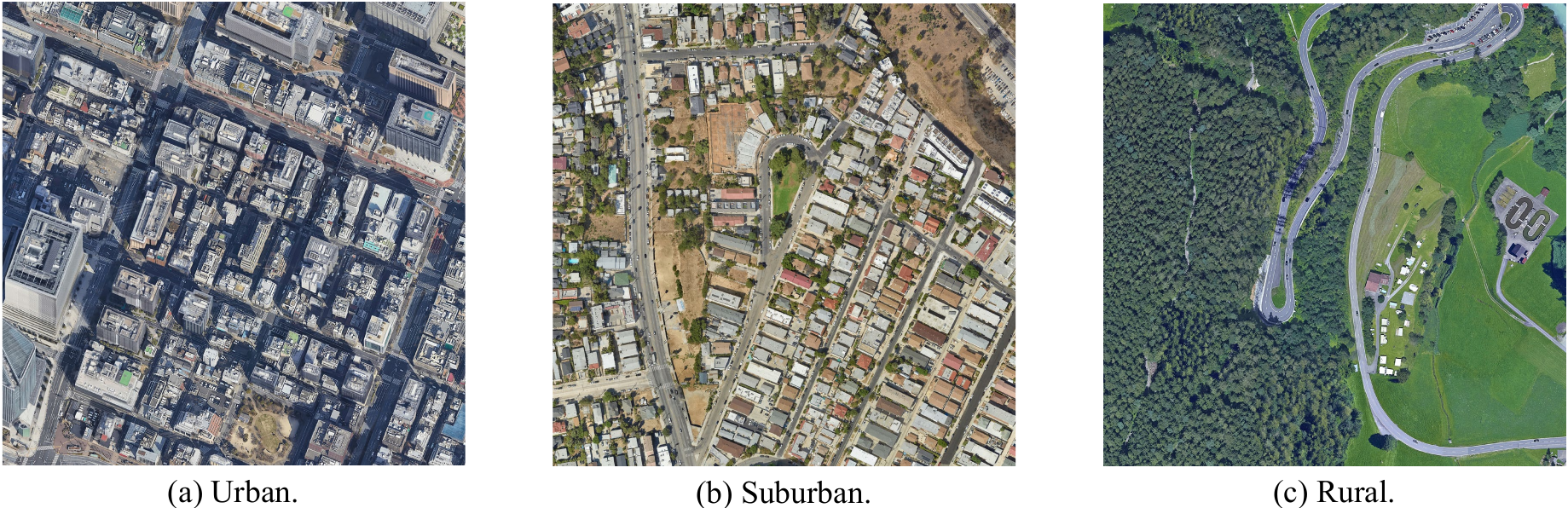}
    \caption{Settlement type examples.}
    \label{fig:area}
\end{figure}

\textbf{Settlement type.} Settlement type is used to describe the surroundings of the road. It includes suburban, rural and urban. 
Figure~\ref{fig:area} shows an example of each case.
\begin{itemize}
    \item Urban: The surroundings are all high rises and office buildings.
    \item Suburban: A form of urban settlement with medium population density, it can be seen as the middle ground between city and country life.
    \item Rural: A human settlement with a low population density. The number of buildings is the least of the three.
\end{itemize}

\begin{figure}
    \centering
    \includegraphics[width=\linewidth]{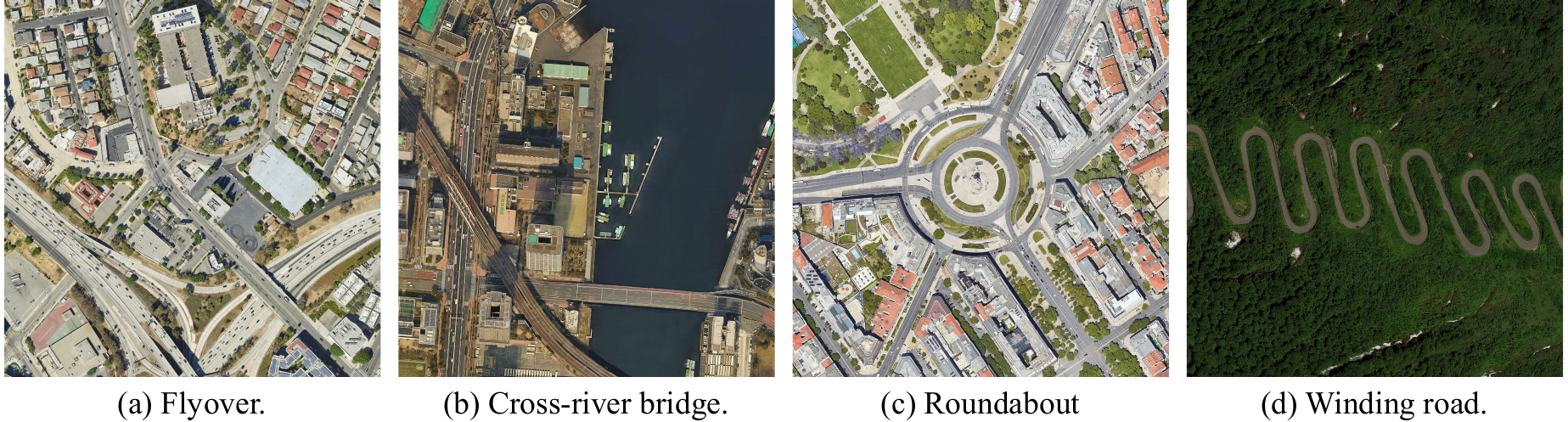}
    \caption{Special road structure examples.}
    \label{fig:structure}
\end{figure}

\textbf{Special road structure.} This tag describes the special road structures of an image and multiple special structures can exist in one image. 
We define four special road structures including roundabout, flyover (overpass), winding road and cross-river bridge.
Figure~\ref{fig:structure} shows an example of each case.

\section{Details of Baseline Methods}
\subsection{Instance-level Line Detection}

\textbf{Implementation details.}
We use MMSegmentation \cite{mmseg2020} to conduct the semantic segmentation.
For \ourdataset19, we cut each image into 512 $\times$  512 and for \ourdataset20, we cut them into 1024 $\times$ 1024.
After cutting process, we resize each image into 2048 $\times$ 2048, \textit{i.e.}, we use "pseudo level-21 images" for segmentation.
In the "pseudo level-21 images", we set the line width as 6 pixel.
We follow the setting of SegNeXt \cite{guo2022segnext} tiny and use cross entroy loss and dice loss as loss function.
All the experiments are conducted on 8 RTX 3090 GPUs.
Table \ref{tab:hyper} shows the hyperparameters we use.

In our baseline method, we use line categories (lane line, curb, background, virtual line) and line types (solid line, thick solid line, dashed line, short dashed line and others) to conduct semantic segmentation and the number of categories is eight.
We get this number according to our definition of instances, \textit{i.e.}, when one attribute changes, we should break the line into two instances. With this definition, we further find only line types and line categories may change between two connected line segments, \textit{e.g.}, a solid line turns into a dashed line or a virtual line breaks the curb. So we only consider these line categories and line types in our semantic segmentation model.
Other attributes do not influence the definition of instances and are not used in this benchmark.

\begin{table}[h]
    \centering
    \vspace{-10pt}
    \setlength{\tabcolsep}{15pt}
    \caption{Hyperparameters in instance-level line detection track.}
    \begin{tabular}{l|l}
    Config & Value  \\
    \shline
    optimizer & AdamW  \\
    learning rate & 8e-5  \\
    weight decay & 0.01 \\
    momentum & $\beta_1$, $\beta_2$ = 0.9, 0.999  \\
    batch size & 8  \\
    learning rate schedule & cosine decay  \\
    warmup iterations & 400   \\
    training iterations & 40000  \\
    augmentation & RandomFlip  \\
    drop path & 0.1 \\
    class weight in CE loss & 1, 20, 30, 40
    \end{tabular}
    \label{tab:hyper}
    \vspace{-10pt}
\end{table}
\subsection{Satellite-enhanced Online Map Construction for Autonomous Driving}
In this subsection, we demonstrate that our dataset could help the online map construction for autonomous driving.
\paragraph{Task definition.}
We follow the setting of semantic map learning, which was first proposed by HDMapNet \cite{li2022hdmapnet}.
Inputs are camera images $\mathcal{I}$ of an autonomous driving vehicle and outputs are vectorized map elements $\mathcal{M}$ around the vehicle.
SatforHDMap \cite{gao2024complementing} adds the corresponding satellite images of each sample as additional input $\mathcal{I_S}\in \mathbb{R}^{ H \times W \times 3}$ and improves the performance, where $H \times W$ indicates the input resolution.
It is a hyperparameter and decided by the region of a satellite map tile (\textit{e.g.}, $60m \times 30m$).
Based on SatforHDMap, we add the mask of the satellite images $\mathcal{I_{MS}} \in \mathbb{R}^{ H \times W }$ as an additional input.

\textbf{Implementation details.}
We use the model from SatforHDMap \cite{gao2024complementing} and add an additional branch to process the mask input.
All the experiments only use the carema images and additional satellite images as input. 
We set the satellite image sampling area as $60m \times 30m$, and the corresponding satellite images tile size is shown in Table~\ref{tab:tile}.
Table~\ref{tab:hypermap} shows the hyperparameters for this track.
\begin{table}[h]
    \centering
    \setlength{\tabcolsep}{15pt}
      \vspace{-10pt}
    \caption{Resolution of satellite map tiles in nuScenes.}
    \begin{tabular}{l|l}
          Region & Resolution \\
           \shline
        Boston Seaport &  545 $ \times $ 273 \\
        Singapore One North & 403 $\times$ 202 \\
        Singapore Queenstown & 403 $\times$ 202\\
        Singapore Holland Village &  404 $\times$ 202
    \end{tabular}
    \label{tab:tile}
    \vspace{-10pt}
\end{table}
\begin{table}[h]
    \centering
    \vspace{-10pt}
    \setlength{\tabcolsep}{15pt}
    \caption{Hyperparameters in satellite-enhanced online map construction  track.}
    \begin{tabular}{l|l}
    Config & Parameters  \\
    \shline
    optimizer & Adam  \\
    learning rate & 0.0016  \\
    weight decay & 1e-7 \\
    momentum & $\beta_1$, $\beta_2$ = 0.9, 0.999  \\
    batch size & 32  \\
    training epochs & 30  \\
    \end{tabular}
    \label{tab:hypermap}
    \vspace{-10pt}
\end{table}

{
    \small
    \bibliographystyle{plain}
    \bibliography{main}
}

\end{document}